\documentclass[11pt,a4paper]{article}
\usepackage{authblk}
\usepackage[hyperref]{emnlp2020}
\usepackage{times}
\usepackage{latexsym}

\usepackage{microtype}

\usepackage{multirow}

\usepackage[T1]{fontenc}
\usepackage[acronym,shortcuts,smallcaps]{glossaries}
\newsavebox{\largestimage}

\usepackage{xcolor}
\usepackage{placeins}
\usepackage{subcaption}
\usepackage{caption}
\usepackage{adjustbox}
\usepackage{enumitem}
\usepackage{placeins}
\usepackage[noend]{algorithm2e}

\setlist[enumerate]{itemsep=0mm, topsep=3pt}
\setlist[itemize]{itemsep=0mm, topsep=3pt}
\newglossaryentry{LSTM}
{
  name={LSTM},
  description={long short-term memory},
  first={\glsentrydesc{LSTM} (\glsentrytext{LSTM})},
  plural={LSTMs},
  descriptionplural={long short-term memories},
  firstplural={\glsentrydescplural{LSTM} (\glsentryplural{LSTM})}
} 

\newglossaryentry{GRU}
{
  name={GRU},
  description={gated recurrent unit},
  first={\glsentrydesc{GRU} (\glsentrytext{GRU})},
  plural={GRUs},
  descriptionplural={gated recurrent units},
  firstplural={\glsentrydescplural{GRU} (\glsentryplural{GRU)}}
}

\newglossaryentry{RNN}
{
  name={RNN},
  description={recurrent neural network},
  first={\glsentrydesc{RNN} (\glsentrytext{RNN})},
  plural={RNNs},
  descriptionplural={recurrent neural networks},
  firstplural={\glsentrydescplural{RNN} (\glsentryplural{RNN})}
} 

\newglossaryentry{NLP}
{
  name={NLP},
  description={natural language processing},
  first={\glsentrydesc{NLP} (\glsentrytext{NLP})},
} 

\newglossaryentry{MT}
{
  name={MT},
  description={machine translation},
  first={\glsentrydesc{MT} (\glsentrytext{MT})},
}

\newglossaryentry{ML}
{
  name={ML},
  description={machine learning},
  first={\glsentrydesc{ML} (\glsentrytext{ML})},
} 

\newglossaryentry{HAN}
{
  name={HAN},
  description={hierarchical attention network},
  plural={HANs},
  descriptionplural={hierarchical attention networks},
  first={\glsentrydesc{HAN} (\glsentrytext{HAN})},
  firstplural={hierarchical attention networks (\glsentryplural{HAN})}
} 

\newglossaryentry{HAN-ST}
{
  name={HAN$_{\textrm{ST}}$},
  description={hierarchical attention network with structure tags},
  descriptionplural={hierarchical attention networks with structure tags},
  first={\glsentrydesc{HAN-ST} (\glsentrytext{HAN-ST})},
  firstplural={hierarchical attention networks with structure tags (\glsentryplural{HAN})}
} 

\newglossaryentry{MSE}
{
  name={MS},
  description={mean squared error},
  first={\glsentrydesc{MSE} (\glsentrytext{MSE})},
} 

\newglossaryentry{MAE}
{
  name={MAE},
  description={mean average error},
  first={\glsentrydesc{MAE} (\glsentrytext{MAE})},
} 

\newcommand\EXCLUDEEMNLPDRAFT[1]{}
\newcommand\EXCLUDE[1]{}

\aclfinalcopy{}

\title{SChuBERT: Scholarly Document Chunks with BERT-encoding boost Citation Count Prediction}

\makeatletter
\newcommand{\removelatexerror}{\let\@latex@error\@gobble}
\makeatother

\author[]{Thomas van Dongen}
\author[]{Gideon Maillette de Buy Wenniger}
\author[]{\textbf{Lambert Schomaker}}
\affil[]{Bernoulli Institute for Mathematics,Computer Science and Artificial Intelligence \protect\\ \small University of Groningen, Groningen, The Netherlands \\
 \texttt{t.a.van.dongen AT student.rug.nl}\\  \texttt{gemdbw AT gmail.com \ l.r.b.schomaker AT rug.nl}}

\date{}

\begin{document}

\maketitle
\begin{abstract}
 Predicting the number of citations of scholarly documents is an upcoming task in scholarly document processing. Besides the intrinsic merit of this information, it also has a wider use as an imperfect proxy for quality which has the advantage of being cheaply available for large volumes of scholarly documents. Previous work has dealt with number of citations prediction with relatively small training data sets, or larger datasets but with short, incomplete input text. In this work we leverage the open access ACL Anthology collection in combination with the Semantic Scholar bibliometric database to create a large corpus of scholarly documents with associated citation information and we propose a new citation prediction model called SChuBERT. In our experiments we compare SChuBERT with several state-of-the-art citation prediction models and show that it outperforms previous methods by a large margin. We also show the merit of using more training data and longer input for number of citations prediction.

\end{abstract}


\section{Introduction}

Predicting the quality of scientific articles is a novel task in the field of deep learning. There are many indicators of quality such as whether a paper was accepted or rejected, meta-information such as the author's h-index(es), and the number of citations. The number of citations, while not a perfect indicator of quality, is available for any paper which makes it suitable for constructing a large dataset. In this work we propose ACL-BiblioMetry, a new dataset consisting of ~30000 papers with citation information.  We also test several state-of-the deep learning models and propose a new model called SChuBERT which outperforms all other methods.

\begin{table*}[t]
\caption{Properties of datasets for citation count prediction applied in earlier work.}
    \centering
    \scalebox{0.89}{
    \begin{tabular}{|c|c|c|c|}
    \hline
         \multirow{2}{*}{paper source} & \multirow{2}{3cm}{\# papers  (train + validation + test)}  &  \multirow{2}{2cm}{\# reviews} & \multirow{2}{*}{paper text type}  \\ 
         & & & \\
         \hline
         \citep{Fu2008} & 3788 & N/A & title+abstract \\
         \citep{li-etal-2019-neural-citation} & 1739, 384 & 7171,  1119  & title + abstract \\
         \citep{PlankAndVanDale2019} & 3427 & 12260  & title + abstract \\
         \multirow{2}{*}{\citep{wenniger2020structuretagsSDP20}} & \multirow{2}{*}{78894 + 4383 + 4382} & \multirow{2}{*}{N/A}  & \multirow{2}{*}{title + abstract + partial body} \\
         & & & \\
         \multirow{1}{*}{ACL-BiblioMetry dataset (this work)} & \multirow{1}{*}{27853 + 1548 + 1549} & \multirow{1}{*}{N/A} & \multirow{1}{*}{title + abstract + full body} \\
         \hline
    \end{tabular}
    }
    
    \label{table:existing-datasets-citation-count-prediction}
\end{table*}

Using the full text of scholarly documents has the potential to substantially improve the performance of the citation count prediction task. But prohibitive memory costs of applying advanced deep learning models on the full text can be a roadblock. In particular, BERT \citep{devlin2018bert} and its variants have been very successful as building blocks for state-of-the-art natural language processing models for many tasks. Citation count prediction for scholarly documents is a task where BERT has clear potential as well. However, scholarly documents are particularly long texts in general.  Since BERT has a time complexity that is quadratic with respect to the input length, it is limited to 512 tokens by default, a limit which can not be increased by much without causing prohibitive computational cost. 

Recent models including the Reformer \cite{kitaev2020reformer} and Longformer \cite{beltagy2020longformer} have sought to overcome the quadratic computational cost of the Transformer model \cite{transformerModel2017}  underlying BERT. While these models are very promising, they do not offer the unsupervised pre-training on large amounts of data that makes BERT so powerful as of yet. Although in principle these models could be applied as a drop-in replacement for BERT, it requires more research to show if and how unsupervised pre-training as done in BERT can be made to work well with very long context. For these reasons, in this work we use BERT as our base building block and find effective ways to overcome its input length limit, leaving experimentation with the aforementioned models for future research. 

For dealing with large amounts of training examples containing very long input text we need an approach that: 1) Is able to fit the encoding of the long text into memory, 2) can efficiently process the large amount of training examples when training over many epochs. Both requirements can be fulfilled by chunking the long input text of our examples into parts, and pre-computing BERT embeddings for each of these parts using a pre-trained BERT model. The core of the final model is a sequence-model, in particular a \ac{GRU} \cite{cho-etal-2014-learning}, which directly uses the pre-computed chunk embeddings as inputs. This approach simultaneously overcomes the memory problems associated with dealing with very long input texts, as well as achieves high computational efficiency by performing the expensive step of computing BERT embeddings for chunks only once.

While the task of citation count prediction using the contents of a scholarly document is not new, and goes back at least to the work of \citet{Fu2008}, work up until now has been limited in: a) the size of the training data,  b) the size of the input text. Table \ref{table:existing-datasets-citation-count-prediction}
gives an overview of data used in earlier work, note that most are restricted by using only the title + abstract as well as a small number of examples, while \cite{wenniger2020structuretagsSDP20} substantially increase the number of examples but still use only a limited part of body text available from S2ORC \cite{s2orc}.
In this work, we show that both these factors have a large influence on the accuracy of models predicting citation counts. Essentially, state-of-the-art methods cannot be adequately evaluated with too small training data.  Therefore, apart from providing state-of-the art results for citation-count-prediction on a data set currently unmatched in terms of number of examples with full length input text, we also provide the code for other researchers to rebuild our dataset and 
the methodology of citation count prediction using the semantic scholar database to label new collections of scholarly documents.

The rest of the paper is organized as follows: in section 2 we discuss related work, in section 3 we describe the models used for citation count prediction, in section 4 we discuss the dataset construction, in section 5 we present our experiments, in section 6 we show our results and in section 7 we end with conclusions.

\section{Related Work}

Recently, multiple datasets have been released which are useful for the scientific quality prediction problem. The S2ORC dataset \cite{s2orc} has abstract information for 81.1M papers and full-text for 8.1M papers, both with citations. 

Other large datasets exist such as unarXive and PubMed Central Open Access Subset, but these datasets span various domains. Given the difficulty of the citation prediction task, we made a new dataset for just the computational linguistics and natural language processing domain, to be used as a benchmark  for citation prediction models.

The PeerRead dataset \cite{kang2018dataset} is another useful dataset that has accept/reject decisions for 14.7K full-text papers. This is a dataset on which more research has been performed \cite{shen2019joint}, but the amount of papers in it is fairly limited. For this reason, we propose our new dataset which contains full-text and citations for a large number of papers. \\
A number of methods have been proposed for the citation prediction problem. \cite{soton260713} try to predict future citations of a paper by using web usage statistics, e.g. the number of times the paper was downloaded. \cite{abrishami} use deep learning techniques to predict long-term citations using short-term citations. \cite{BAI2019407} use a measure called Paper Potential Index (PPI) which is based on a combination of features such as the impact of the authors and early citations.
The problem with these methods is that information such as short-term citations and web usage statistics are only available after the paper is published. Furthermore, these methods disregard any of the actual papers' content. 

Because of this reason, our work focuses on predicting the citations using only the textual content. Limited research is available on this topic. One of the first papers which focused on predicting citation count by only using information available at publication is by \cite{Fu2008}. They use the paper title, abstract and keywords as well as bibliometric information as input data for an SVM. They then predict a binary label (positive or negative) based on whether the paper received at least a set number of citations within 10 years. This work was expanded upon by \cite{10.1093/bioinformatics/btp585}. They predict a discrete value (few, some or many citations) using multiple classification models, outperforming the baseline set by \cite{Fu2008} using both naive Bayes as well as logistic regression. Both papers use a fairly small dataset (3788 papers for \cite{Fu2008} and 2246 papers for \cite{10.1093/bioinformatics/btp585}).

The task of predicting sequences of citation counts, was recently proposed by \citet{holm2020longitudinal}. They use 
citation network information, in addition to paper abstract, author and venue information. 
Their work distinguishes itself from previous work in the field, by combining topological and temporal information from citation networks,  using graph convolution networks paired with sequence prediction.
Notably, though, their proposed task and ours differ significantly. We focus on the task of citation count prediction as a proxy of quality prediction and therefore deliberately use only textual content, omitting explicit or implicit quality labels in the input, including author and venue; as we argue these may obscure the role of the paper content in the prediction. In contrast, \citet{holm2020longitudinal} focus on the sequence prediction aspect, and include quality-indicative labels (author, venue), which changes the nature of the task substantially.

\section{Models}

In this section we briefly describe our two baseline models: BiLSTM and \acp{HAN}. This is followed by 
a description of the BERT-based SChuBERT model, to the best of our knowledge first applied to the task of citation count prediction in this work.

\subsection{BiLSTM Based Prediction}

\EXCLUDE{
\begin{figure*}
    \centering
    \scalebox{0.5}{
    \includegraphics{BiLSTM-model-color-uml.pdf}
    }
    \caption{The BiLSTM baseline model used in this work. Reproduced from \cite{wenniger2020structuretagsSDP20}.}
    \label{fig:bilstm-model}
\end{figure*}
}

Our BiLSTM baseline model, is a re-implementation of the BiLSTM model introduced in \cite{shen2017hybrid} . This model was initially used for the task of Wikipedia text quality prediction and applied also in \cite{shen2019joint} for the task of accept/reject prediction on the PeerRead dataset, and finally in \cite{wenniger2020structuretagsSDP20} for the task of citation count prediction. The name ``BiLSTM'' is somewhat deceptive as the model contains several other layers in addition to a plain BiLSTM to improve performance: 
\begin{enumerate}
    \item The sentence embeddings in the input are fed to an average pooling layer, to combine them to a single representation per input sentence.
    \item Following the BiLSTM is a max-pooling layer followed by a rectified linear hidden layer. These additional layers are added to further improve performance. 
\end{enumerate}

The simplicity of the sentence encoding employed by this model yields relatively high computational efficiency, lower memory usage and scalability to longer input text. This makes the model competitive in settings where the amount of training material is limited, such as PeerRead accept/reject prediction
\cite{shen2019joint,wenniger2020structuretagsSDP20}. However, as we will show later in this work, there is a clear advantage to using the more advanced SChuBERT model given enough training data is available.

\subsection{Hierarchical Attention Networks}

\begin{figure*}[t]
    \centering
    \scalebox{0.5}{
    \includegraphics{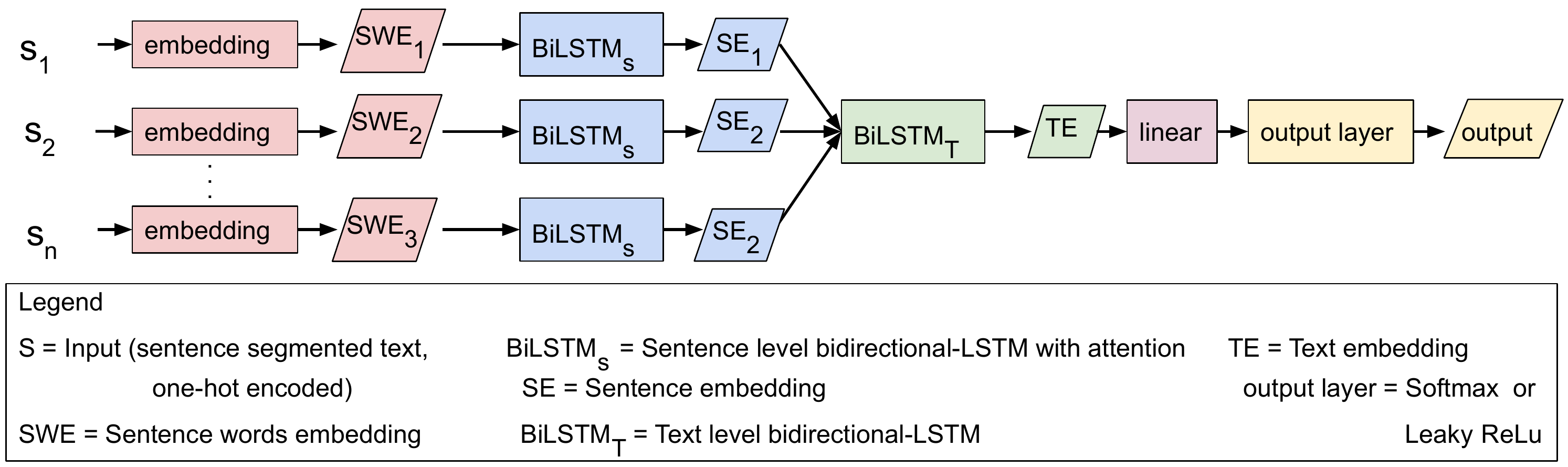}
    }
    \caption{The HAN baseline model used in this work. Adapted from \cite{wenniger2020structuretagsSDP20}.}
    \label{fig:han-model}
\end{figure*}

The \ac{HAN} model \cite{han2016}, see Figure \ref{fig:han-model}, used in this work is a PyTorch re-implementation of the original model.\footnote{Adapted from https://github.com/cedias/Hierarchical-Sentiment}
It is in some ways similar to the BiLSTM model discussed earlier, but creates more advanced sentence-level representations by applying a BiLSTM with attention for encoding these as well as employing a BiLSTM with attention for for converting the sentence-level representations to document-level representations.

We next discuss the more advanced BERT-based model.

\subsection{SChuBERT}

\begin{figure*}[t]
    \centering
    \scalebox{0.6}{
    \includegraphics{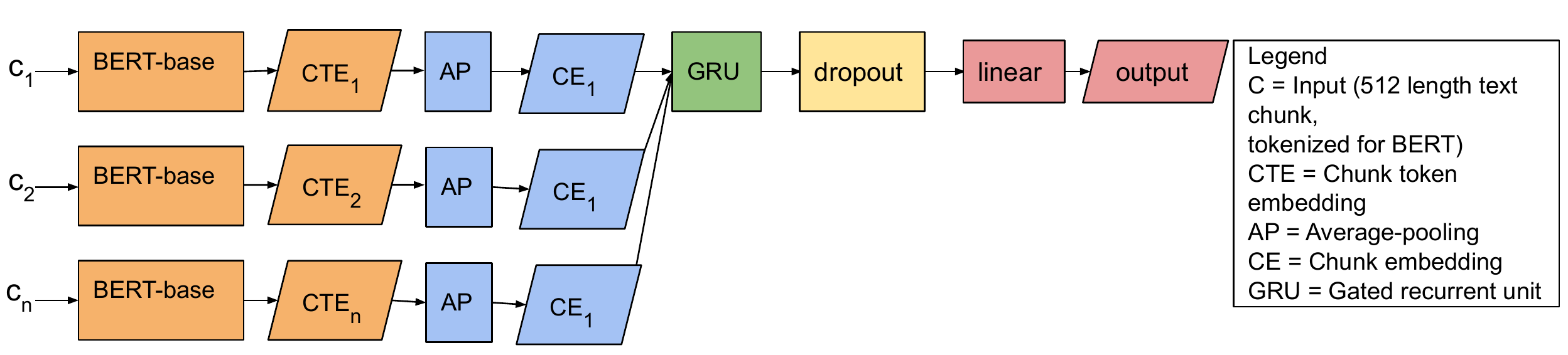}
    }
    \caption{The SChuBERT model proposed in this work.}
    \label{fig:schubert-model}
\end{figure*}

Our SChuBERT model, shown in fig \ref{fig:schubert-model}, consists of two parts: a pre-trained BERT (Bidirectional Encoder Representations from Transformers)  model \cite{devlin2018bert} to extract features and a deep learning model to learn from the features and predict. The main difference between this model and the other models is the use of contextualized word embeddings instead of context-independent word embeddings such as the Glove embeddings used in the HAN model. These offer a much richer context by not just encoding a word using a static embedding but encoding it based on the context it appears in. 

One limitation of transformer-based models such as BERT is that they have a time complexity of $\mathcal{O}(N^2)$ with respect to the input length. For this reason, most of these models are pre-trained on sequences of a maximum length of 512. Since we are dealing with very long sequences, we have to work around this limit. The simplest approach is to truncate the documents to a length of 512 as proposed in \cite{xie2019unsupervised}. However, since our documents are so long, this would remove a lot of information. For this reason, we adopt the technique proposed in \cite{m2019bert}. We split each input into chunks of 512 with an overlap of 50 tokens each to preserve a relation between the chunks. 

For pre-trained BERT models for feature extraction, there are two considerations to make. Firstly, since BERT generates an embedding of length 768 for each token in our chunks of (max) 512 tokens, we need to pool over these embeddings to get embeddings of equal length. Note that the CLS (classification) token, which is normally used for classification tasks, is not a good representation without fine-tuning since it only holds useful information for the pre-training tasks when no fine-tuning is performed on the target domain. For this reason, we use mean pooling over our embeddings. 
Secondly, the different layers in BERT hold different information. The earlier layers are closer to the original word embeddings, which in the case of BERT are WordPiece embeddings \cite{wu2016googles}, while the later layers are closer to the pre-training targets. Intuitively, it would make sense that the last layers are too close to the pre-training targets and are thus biased. However, our findings correspond with \cite{peters2019tune} in which the last layer (layer 12) is found to be the most useful for feature extraction which is why we use this layer.

After extracting the embeddings, they are passed through a fairly simple model to do predictions. We use a GRU, followed by a single dropout layer and a linear layer.
We use a simple model since the embeddings already hold a lot of information and are prone to over-fitting when a more complex model is used.

\begin{figure*}[t!]
\footnotesize

\{"entities":[],"journalVolume":"","journalPages":"97-115","pmid":"","fieldsOfStudy":["Computer Science"], \\
"year":2019,\textbf{"outCitations":
["c91f19447f7a72afe58ecf7281033df276b20497",\\
"bd59f9543127f56074aa2e6adb259099eb333912", "acbd8a36a59b7e27ddf24b64133b6b9cf4c6990c",\\ "d8c1b48ae4d6e4676d060c06087bb6b1ac81a005", 
\ldots ,"2671a510c47b7fbe117fa07051829914cd1b4c98"]},\\
"s2Url":"https://semanticscholar.org/\\paper/3958cfb18ce6f32e90bd6ef5473be7ddd5a4e464",
"s2PdfUrl":"", \textbf{"id":"3958cfb18ce6f32e90bd6ef5473be7ddd5a4e464"},\\
"authors":[\{"name":"Tim van de Kamp","ids":["7401984"]\},\{"name":"David Stritzl","ids":["146553639"]\},\{"name":"Willem Jonker","ids":["6235263"]\},\{"name":"Andreas Peter","ids":["144253636"]\}],"journalName":"", \\
"paperAbstract":"We propose several functional encryption schemes for set intersection and variants on two or multiple sets.  \\

\ldots \\

\textbf{"inCitations":["f1a2ab3038bedbdfabd35f8d41103b99f51d0ec7"]},
"title":"Two-Client and Multi-client Functional Encryption for Set Intersection","doi":"10.1007/978-3-030-21548-4\_6","sources":["DBLP"],"doiUrl":"https://doi.org/10.1007/978-3-030-21548-4\_6","venue":"ACISP"\}
 \caption{Example of a JSON paper entry in the Semantic Scholar database source files. The paper id, outCitations,  and inCitations are shown in bold, for clarity.}
 \label{figure:semantic-scholar-database-entry-example}
\end{figure*}


\section{Dataset construction}

Citation count prediction relies on sufficiently large labeled data, of good quality and preferably with 
full document text. To obtain such data, we need:
\begin{enumerate}
    \item A large set of good quality scholarly documents, preferably in the same domain, or a way to collect such a set from the internet.
    \item A scalable way to obtain citation counts for papers , and a way to restrict the citation counting to a fixed number of years after a paper's publication, in order to get comparable counts for papers that are published in different years. 
\end{enumerate}

To accomplish these, we first discuss a method to collect papers from the ACL Anthology database, yielding a relatively large set of full text documents, of notably good quality and relatively controlled length in comparison to other alternatives such as the arXiv repository which we also considered. The resulting dataset is called ACL-BiblioMetry.\footnote{Link to scraper code and citation information data: https://github.com/Pringled/ACL-BiblioMetry} We next discuss a method to collect the required citation counts.

\begin{figure*}
    \begin{minipage}{.52\textwidth}
        \begingroup
    \removelatexerror
          \begin{algorithm*}[H]{scrapeACL}
 \SetAlgoLined\\
  \KwData{None.}
  \KwResult{A folder of PDF and corresponding bib files.}
  
  relevant\_links $\leftarrow$ extractLinks(page) \\
  \For{link $\in$ relevant\_links}
  {
    title $\leftarrow$ getVenueAndYear(link); \\
    page\_source $\leftarrow$ getPageContent(link); \\
    \For{line $\in$ page\_source}
    {
        \If{in\_line("pdf") and not (in\_line("poster") \\ or  in\_line("presentation") or\\
        in\_line("supplementary") or in\_line("notes"))}
		{   
		    pdf\_link  $\leftarrow$ extractPdfLink(line) \\
		    bib\_link  $\leftarrow$ extractBibLink(line) \\
		    downloadPdf(pdf\_link) \\
		    downloadBib(bib\_link)
		}        
    }
    
  }
  \Return None \\ 

  \caption{Algorithm for scraping the ACL anthology database.}
\vspace{0.2cm}
\label{Algorithm:scrapeACL}
\end{algorithm*} 
\endgroup
    \end{minipage}
    \begin{minipage}{.48\textwidth}
        \begingroup
    \removelatexerror
   \begin{algorithm*}[H]{findCitationsForArticleFromDatabase}
 \SetAlgoLined\\
  \KwData{$\langle$authors\_table, articles\_table$\rangle$, $\langle$title: String, authors: list$\rangle$.}
  \KwResult{A dictionary of $\langle$year, citation\_ids\_list$\rangle$ entries.}
  
  \For{author $\in$ authors}
  {
    article\_ids $\leftarrow$ selectIDsWithAuthor(authors\_table, author); \\
    \For{article\_id $\in$ article\_ids}
    {
        article $\leftarrow$ selectArticleWithID(articles\_table, article\_id); \\
        \If{article.title  = title}
		{
            \Return computeYearGroupedCitations(article)
		}        
    }
    
  }
  \Return None \\ 
  
\vspace{0.2cm}

\caption{Algorithm for matching an article title and authors list to the database, returning the citations information for the 
first found article that matches the title and one of the authors.}
\label{Algorithm:findCitationsForArticleFromDatabase}
\end{algorithm*} 
\endgroup

\end{minipage}
\end{figure*}

\subsection{Scraping ACL}
For retrieving the data from the ACL Anthology database, we use the method described in Algorithm \ref{Algorithm:scrapeACL}. First, all relevant links are extracted from the ACL Anthology main page. This includes links to all listed venues from all years. The venue as well as the year is saved for each entry as they are used as names for the saved PDF and bib files. Then, for each link, the page source is retrieved. In the page source, all relevant links to PDFs are found. PDF links that correspond to posters, presentations, supplementary materials and notes are ignored. After this, the bib link corresponding to the PDF link is extracted and both are saved using the venue name and year.

\subsection{Citations Retrieval}

To retrieve the set of papers that cites a given paper, we use the Semantic Scholar database.\footnote{The Scholar Database source files are available from  https://api.semanticscholar.org/corpus/download/ .} 
The database in its provided form consists of a collection of JSON objects, one per line. Figure \ref{figure:semantic-scholar-database-entry-example} shows an example of an entry from the database source-files.
Each entry has an \emph{id}, a list of paper \emph{outCitations}: IDs of the papers that the entry paper cites, as well as a list of paper \emph{inCitations}: IDs of papers that are citing the entry paper. For our purposes in this work we are mainly interested in the \emph{inCitations} information. Naively, the raw semantic scholar database entries already provide us with the information of how often a paper is cited. However, in practice this is not very useful, since papers are published in different years. Consequently, more recent papers will have had much less time to ``collect'' citations. To correct for this, and get comparable citation counts, we need to count only citations within a fixed window of time from each paper's data of publication. The latter task is slightly more involved to solve, noting that every source file is 1.6 Gigabytes, with 185 source files for a total of 283 Gigabytes of text data at the time of writing.

\subsection{SQL database for efficient retrieval}

By creating an SQL database that contains all the information in a structured way with proper indices, the task becomes manageable.
Specifically we create a database consisting of two tables:
\begin{enumerate}[topsep=0pt,itemsep=-1ex,partopsep=1ex,parsep=1ex]
 \item Authors table. Fields: [article\_id (text, PRIMARY KEY), author\_name (text).]  \\
 An index is added to author\_name, to facilitate fast lookup of papers that have a certain author.
 \item Articles table. Fields: 
  [article\_id (text. PRIMARY KEY), title (text), pages (text), year (text),
   volume (text), journal (text), inbound\_citations (text), outbound\_citations (text),  doi (text)].  \\
   The fields in the articles table are kept quite minimal, omitting some unnecessary information from the original semantic scholar source files.
   An index is added to article\_id for fast lookup of a paper with a given article\_id.
\end{enumerate}

\begin{table*}[t]
\centering
\caption{Character count per example statistics ACL dataset different settings.}
\scalebox{0.95}{
    \begin{tabular}{|c|c|c|c|c|c|}
    \hline
     systems & \multirow{2}{2.5cm}{BiLSTM, HAN, SChuBERT} & 
     \multirow{2}{1.5cm}{BiLSTM, HAN} & \multicolumn{3}{c|}{\multirow{2}{*}{SChuBERT}}  \\
     & & & \multicolumn{3}{c|}{} \\
    \hline
    setting &  \multirow{3}{3.5cm}{title + abstract} & \multirow{3}{3cm}{title + abstract + body text (max 200000 chars)} & \multicolumn{3}{|c|}{title + abstract + body text}  
      \\
    & & &\multirow{2}{2cm}{max 5 chunks} & \multirow{2}{2cm}{max 6 chunks} & \multirow{2}{1.5cm}{no limit} \\
    & & & & & \\
    \hline
    \multirow{2}{2cm}{\#characters
    (avg, max)} & \multirow{2}{2cm}{975 , 20000} &  \multirow{2}{2.5cm}{17293 , 20000} & 
    \multirow{2}{1.5cm}{12019 , 19064} &\multirow{2}{1.5cm}{14061 , 22643}  & \multirow{2}{1.5cm}{23787 , 1261656} \\
    & & & & & \\
    \hline 
    \end{tabular}
    \label{table:character-counts-per-example-different-settings}
}

\end{table*}

\subsubsection{Database creation}

After creating the database the two tables are filled by simply looping over the semantic scholar source files and adding a corresponding entry to the articles table for each article entry in the source files. The authors table in addition is filled with an entry for each author of the article entry. The aim of this is that an article can be retrieved based on each of the author's names separately, increasing recall. To further increase recall all the author names are lowercased (in the created database and during retrieval).

 \begin{algorithm}[t]{computeYearGroupedCitations}
 \SetAlgoLined\\
  \KwData{$\langle$authors\_table, articles\_table$\rangle$, article.}
  \KwResult{A dictionary of $\langle$year, citation\_ids\_list$\rangle$ entries.}
  
  result\_dict $\leftarrow$ dict([]);
  
  \For{citing\_article\_id $\in$ article.in\_citations}
  { 
    citing\_article $\leftarrow$ selectArticleWithID(articles\_table, citing\_article\_id); \\
     \If{not(citing\_article.year  $\in$ result\_dict)}
     {
        result\_dict[year] $\leftarrow$ list([]);
     }
     result\_dict[year].\\
     \  \ \ \ append(citing\_article.article\_id);
    
  }
  \Return result\_dict  \\ 
  
\vspace{0.2cm}

\caption{Algorithm for generating a dictionary of ids of citing articles, collected in sub-lists indexed by year.}
\label{Algorithm:computeYearGroupedCitations}
\end{algorithm}

\subsubsection{Number of citations retrieval}

Given an article, the citations of the article are retrieved from the database based on the authors list and title of the paper. 
This is done in two stages, shown also in Algorithm \ref{Algorithm:findCitationsForArticleFromDatabase}:
\begin{enumerate}
 \item Paper retrieval: One by one, for each of the authors, all paper ids are retrieved. From these, matching article entries are found from the articles table. The first paper by any of the authors that matches the query title is returned as a positive match.\footnote{We require only one author name to match, because this significantly increases recall, while the chance of false positives given the full title and one fully matching author name is negligible.} Just as the case for author names, titles are also lowercased to further increase recall.
 \item Once the correct article entry is retrieved, the list of paper IDs of inbound citations, can be obtained from this entry. 
 For each of these IDs an article entry is obtained and from that entry the publication year of that article. Finally, the IDs of the citing papers are grouped in a dictionary indexed by year
 (see Algorithm \ref{Algorithm:computeYearGroupedCitations}).

\end{enumerate}

\begin{table}
\caption{Hyperparameters used in the experiments.}
\centering
\scalebox{0.80}{
 \begin{tabular}[H]{|l|l|l|}
\hline
                          & \multicolumn{1}{c|}{\multirow{2}{2cm}{BiLSTM and HAN}}    & \multirow{2}{1.0cm}{SChuBERT}       \\
&  &      \\                          
\hline                          

vocabulary size &  \multicolumn{2}{|c|}{10000} \\   
weight initialization    &   \multicolumn{2}{|c|}{} \\                                                  
              \multicolumn{1}{|r|}{general} &  \multicolumn{2}{|c|}{Xavier uniform} \\                                      
              \multicolumn{1}{|r|}{lstm}    &  \multicolumn{2}{|c|}{Xavier normal} \\                                     
              \multicolumn{1}{|r|}{bias}    &  \multicolumn{2}{|c|}{zero}         \\                                       
              
              \multirow{2}{4cm}{optimizer, learning rate}               & \multirow{2}{1.5cm}{Adam, 0.005}   & \multirow{2}{1.5cm}{Adam, 0.001} \\
              & & \\
              epochs              & \multicolumn{1}{|l|}{160}   & 30 \\
              maximum input characters &  \multicolumn{1}{|l|}{20000} & no limit \\                                              
              \multicolumn{1}{|l|}{word embeddings}    &  \multicolumn{1}{|l|}{GloVe}         & \multicolumn{1}{|l|}{N/A}         \\
              \multirow{1}{*}{loss function}            & \multirow{1}{*}{MAE}              & \multirow{1}{1.7cm}{MAE} 
              \\
dropout probability      & 0.5                        & 0.3                   \\
BiLSTM/GRU hidden size       & 192                        & 512                   \\
batch size               & 4, 16                          & 12                    \\
word embedding size           & 50                         & N/A                   \\
BERT sentence embedding size           & N/A                        & 768                   \\
\hline
  
 \end{tabular}
}

\label{table:hyperparameters}
\end{table}

\begin{table*}[t]
\center
\caption{Results on the full data and with full input.}
\scalebox{0.9}{
 \begin{tabular}{|c|c|c|c|c|c|}
 \hline 
   & BiLSTM & HAN & SChuBERT (5 chunk) & SChuBERT (6 chunk) & SChuBERT \\
   \hline
$R^2$ score & 0.319  $\pm$ 0.013 & 0.339 $\pm$ 0.013 & 0.369 $\pm$ 0.009 & 0.380 $\pm$ 0.004  & \textbf{0.398 $\pm$  0.006}\\   
\hline 
MSE & 1.110 $\pm$  0.021 & 1.080 $\pm$  0.021 & 1.032 $\pm$ 0.015 & 1.013 $\pm$ 0.006 &   \textbf{0.985 $\pm$   0.010}\\
\hline
MAE & 0.824 $\pm$ 0.009  & 0.820 $\pm$ 0.009 & 0.805 $\pm$ 0.005 & 0.798 $\pm$ 0.005 & \textbf{0.789 $\pm$ 0.005} \\
 \hline  
\end{tabular}
}

\label{table:main-system-comparison-results}
\end{table*}

\begin{table*}[t]
\begin{minipage}[t]{0.55\textwidth}
\caption{Results on the full data and with abstract text only.}
\center
\scalebox{0.9}{
 \begin{tabular}{|c|c|c|c|}
 \hline 
   & BiLSTM & HAN & SChuBERT \\
   \hline
$R^2$ score & 0.158   $\pm$ 0.006 &  0.248 $\pm$ 0.014  & \textbf{0.249 $\pm$  0.002}\\   
\hline 
MSE & 1.377 $\pm$ 0.010  & \textbf{1.230 $\pm$ 0.023}  &  \textbf{1.230 $\pm$   0.004}\\
\hline
MAE & 0.933 $\pm$ 0.002  & 0.885 $\pm$ 0.008  & \textbf{0.884 $\pm$ 0.002} \\
 \hline  
\end{tabular}
}

\label{table:main-system-comparison-results-abstract}
\end{minipage}
\begin{minipage}[t]{0.4\textwidth}

\center
\caption{Results for SChuBERT on a subset of the data and with full input.}
\scalebox{0.9}{
 \begin{tabular}{|c|c|c|c|}
 \hline 
   &   \multirow{2}{2cm}{SChuBERT 50\% data} &   \multirow{2}{2cm}{SChuBERT 10\% data} \\
   & & \\
   \hline
$R^2$ score & 0.327  $\pm$  0.007 & 0.205 $\pm$  0.026 \\   
\hline 
MSE & 1.058 $\pm$  0.011 & 1.473 $\pm$  0.048 \\
\hline
MAE & 0.809 $\pm$ 0.005  & 0.923 $\pm$  0.027 \\
 \hline  
\end{tabular}
}
\label{table:system-comparison-results-subset-data}

\end{minipage}
\end{table*}

\begin{table}[ht]
\caption{Number of trainable parameters for used hidden sizes.}
\begin{tabular}{|l|l|l|l|}

\hline
Hidden size & BiLSTM  & HAN     & SchuBERT \\ \hline
192         & 1170949 & 2059525 &    N/A      \\ \hline
256         & N/A        &  N/A       & 788225   \\ \hline
512         & N/A        &  N/A       & 969665   \\ \hline
\end{tabular}
\label{table:parameters}
\end{table}

\begin{table}[ht]
\caption{Training time per epoch in seconds.}
\begin{tabular}{|l|l|l|l|}
\hline
                & BiLSTM & HAN  & SchuBERT \\ \hline
Time in seconds & 1048   & 1921 & 12       \\ \hline
\end{tabular}
\label{table:time-per-epoch}
\end{table}

\begin{table}[ht]
\caption{ Results for SChuBERT with hidden size 256 (with full data).}
\scalebox{0.95}{
\begin{tabular}{|l|l|l|}
\hline
MSE           & MAE           & R2            \\ \hline
0.994 $\pm$ 0.013 & 0.788 $\pm$ 0.006 & 0.392 $\pm$ 0.008 \\ \hline
\end{tabular}
}
\label{table:SChuBERT-hidden-size-256}
\end{table}

\subsubsection{Citation scores and year-range uniformity}

In our work we follow \citet{wenniger2020structuretagsSDP20} in using citation scores defined as 
\begin{equation}
 \textrm{citation\_score} = log(\textrm{number\_of\_citations} + 1)
\end{equation} 
When computing these (or other) scores, it is critical to use uniform year-ranges, that is a uniform $\textrm{MAX\_YEARS}$: the maximum years after the publication of an article for collecting citations. A secondary question is: what are good values for 
$\textrm{MAX\_YEARS}$?
We believe in principle higher values will reduce the effects of randomness in the scores, and therefore it seems reasonable to allow at least a few years (e.g. setting $\textrm{MAX\_YEARS} > 3)$. Taking this into account, we believe that whereas enforcing $\textrm{MAX\_YEARS}$ uniformly is important, the value chosen for it is less important: all large enough values will give citation\_score distributions such that the citation\_score can be used (to some extent) to reflect the relative quality or impact of articles. Therefore, we leave finding an optimal value for future work. Even so, the advantage of the way we collect the citation information is that it is straightforward to experiment with different settings.
One practical reason however for not choosing the parameter too large is that it disallows more recent publications to be included. For example, at the time of writing (August 2020) setting $\textrm{MAX\_YEARS}$ to 3 means that articles published up to 2016 can be included, as they have 3 complete years after 2016 (i.e. 2017, 2018, 2019) to ``collect'' citations. Papers published after 2016 cannot be included with this setting. We used this setting in our experiments,
as we believe it to be large enough to give reliable citation\_score values, while small enough to allow inclusion of a large number of articles in the data.\\ \\

\noindent{\textbf{Computation}} \\ 
Once  $\textrm{MAX\_YEARS}$  is chosen, for an article $a$, and and associated citations dictionary $a\_citations\_dict$ computed by Algorithm \ref{Algorithm:computeYearGroupedCitations} selecting included citations is easy. Simply concatenate the lists of year-indexed citations sublists with $year(sublist) \leq a.year + \textit{MAX\_YEARS}$.
Based on the final list of included citations, the citation score or other metrics can then be easily computed.

\section{Experiments}

In our experiments, we want to assess the usability of the ACL data for number of citations prediction, and generally larger training data for citation prediction, as enabled by the automatic number of citation labeling framework contributed in this work. We also want to test two hypotheses:
\begin{enumerate}[topsep=0.5pt,itemsep=0pt,partopsep=1ex,parsep=1ex]
 \item Longer input text improves performance: using then entire paper text (title + abstract + body text) is substantially better than using only the paper title + abstract.
 \item Larger training data substantially improves performance. More specifically,  when using training data for number of citation prediction that is $n$ times larger than what has been used for the related task of accept/reject prediction on the PeerRead CL dataset (computation and language domain) yields substantially better results than when using a training set of size comparable to PeerRead CL.
\end{enumerate}

To test these two hypothesis, we perform the following comparisons:

\begin{enumerate}
    \item Full text input in comparison to abstract only.
    \item Full data input in comparison to 50\% data input and to 10\% data input.
\end{enumerate}

To test our second hypothesis, we take 10\% of our dataset to get a dataset which is approximately the size of PeerRead CL (~3000 papers). We then compare this to half our data and full data to show the importance of larger datasets. Lastly, we also test SChuBERT on a portion of the chunks to ensure a fair comparison with BiLSTM and HAN which were capped at 20k characters. 
Statistics about the number of characters per example in the different settings
are shown in Table \ref{table:character-counts-per-example-different-settings}.

\subsection{Experimental Settings} 
Table \ref{table:hyperparameters} shows the hyperparameters used for training the models in our experiments. 
As evaluation metrics, we report the standard metrics of \ac{MSE} and \ac{MAE}, which are commonly used for regression evaluation, as well as the $R^2$ score.
We repeat each experiment three times to counter false conclusions due to optimizer instability, and report average and standard deviation for each of the metrics.

\section{Results}
Our results show that SChuBERT is able to outperform both BiLSTM as well as HAN for the citation prediction problem by a significant margin. Table \ref{table:main-system-comparison-results} shows a comparison of the three models for full data input and full-text input.  While BiLSTM and HAN have a comparable $R^2$ score, SChuBERT has an $R^2$ score of almost 0.06 higher, showing the power of contextualized word embeddings. SChuBERT also performs better with 5 chunks (which equates to less total input used than BiLSTM and HAN which were capped at 20k characters) and 6 chunks (which equates to slightly more input used than BiLSTM and HAN). For reference, the average number of chunks was 7.6. In practice, capping the chunks mostly results in extremely long papers being cut off, just like in BiLSTM and HAN.

We also compared how well the different models performed on abstract-only inputs, shown 
in Table \ref{table:main-system-comparison-results-abstract}. These results show that HAN and SChuBERT have comparable results, which indicates that SChuBERT benefits more from longer inputs. In general, the performance of all models is substantially better on full-text inputs when compared to abstract only. 

In Table \ref{table:system-comparison-results-subset-data} we show the performance of SChuBERT on less data. As expected, the performance decreases substantially with less data, showing the benefit of larger datasets such as the one proposed in this paper. Due to time constraints, we did not test BiLSTM and HAN on less data. 

As a final comparison of the systems, we show the number of trainable parameters in Table \ref{table:parameters} and the training time in seconds per epoch in Table \ref{table:time-per-epoch}. As can be seen, SChuBERT has a smaller number of trainable parameters even with a larger hidden size for the GRU. Additionally, in Table \ref{table:SChuBERT-hidden-size-256} we show results for SChuBERT when we half the hidden size to 256, which turns out to only give a small drop in performance. The training time in seconds per epoch is also much lower for SChuBERT, which trains approximately 87 faster than BiLSTM and 160 times faster than HAN. However, this is after the embeddings have been generated, which takes relatively long (a little over 7 hours for the full dataset) but only has to be done once. Even when taking this into consideration, training SChuBERT is still much faster given that it converges about 4 times faster than the other systems.

\section{Conclusion}
In this work, we showed the importance of larger and better curated data for the citation prediction problem. We proposed ACL-BiblioMetry, a new large dataset created with the algorithms we provide in this work. We also proposed SChuBERT, a new model for the citation prediction problem which can deal with large inputs and gets significantly better results than several state-of-the-art models. The model shows the strength of modern language models and contextualized word embeddings and their appliance to the citation prediction problem. Our results indicated that both the length of the input as well as the amount of data are important for achieving better results. 
The current work takes a step forward by using a larger training set of full text examples and leveraging this data with stronger models, in particular the SChuBERT model, without considering the historical publication context and other factors. We leave experimentation with further extended context for the predictive models, as well as other language models and even larger datasets for future work.

\section*{Acknowledgments}
The Peregrine high performance computing cluster, at the Center for Information Technology of the University (CIT) of Groningen, was used for running part of the experiments in this study.
We would like to thank the people  at the CIT for their support and access to the cluster and the anonymous reviewers for comments and suggestions. We would also like to thank Charles-Emmanuel Dias for sharing his HAN implementation, which proved to be a solid foundation for the HAN models used in this work.

\bibliography{references}
\bibliographystyle{acl_natbib}

\end{document}